\documentclass[11pt]{article}

\usepackage[preprint]{acl}
\usepackage{amsmath}

\usepackage{times}
\usepackage{latexsym}

\usepackage[T1]{fontenc}
\usepackage[utf8]{inputenc}

\usepackage{microtype}

\usepackage{inconsolata}
\usepackage{tabularx}
\usepackage{graphicx}
\usepackage{enumitem}
\usepackage{booktabs}
\usepackage{listings}
\usepackage{xcolor}

\title{Multi-Agent Comedy Club: Investigating Community Discussion Effects on LLM Humor Generation}

 \author{
 	\textbf{Shiwei Hong\textsuperscript{1}},
 	\textbf{Lingyao Li\textsuperscript{2}},
 	\textbf{Ethan Z. Rong\textsuperscript{3}},
 	\textbf{Chenxinran Shen\textsuperscript{3}},
 	\textbf{Zhicong Lu\textsuperscript{1}} 	
 	\\
 	\\
 	 \textsuperscript{1}George Mason University,
 	 \textsuperscript{2}University of South Florida,
 	 \textsuperscript{3}University of Toronto
 	 \\
 	 \\
 	        \UrlFont{ \{shong46, zlu6\}@gmu.edu}
 	}

\begin{document}
\maketitle
\begin{abstract}
Prior work has explored multi-turn interaction and feedback for LLM writing, but evaluations still largely center on prompts and localized feedback, leaving persistent \emph{public} reception in online communities underexamined.
We investigate whether broadcast community discussion improves stand-up comedy writing in a controlled multi-agent sandbox: in the discussion condition, critic/audience threads are recorded, filtered, stored as social memory, and later retrieved to condition subsequent generations, whereas the baseline omits discussion.
Across 50 rounds (250 paired monologues) judged by five expert annotators using A/B preference and a 15-item rubric, discussion wins 75.6\% of instances and improves Craft/Clarity ($\Delta{=}0.440$) and Social Response ($\Delta{=}0.422$), with occasional increases in aggressive humor.
\end{abstract}

\section{Introduction}

Large Language Models (LLMs) are increasingly deployed as writing assistants that personalize outputs using authors' historical documents and generate actionable feedback for revision \citep{mysore-etal-2024-pearl, chamoun-etal-2024-automated}. Given that LLM-generated texts are now prevalent on social media and significantly impact human readers \citep{radivojevic2024humanperception}, a natural hypothesis arises: networked discussions may, in turn, influence LLMs. This explicitly mirrors core dynamics in online communities, where creative writing is inherently linked to public reception (e.g., comments and critiques) that authors use to refine their work \citep{guo2023onlinecritiques,cheng2022fanfiction}.

\begin{figure*}[t]
  \centering
  \includegraphics[width=0.95\textwidth]{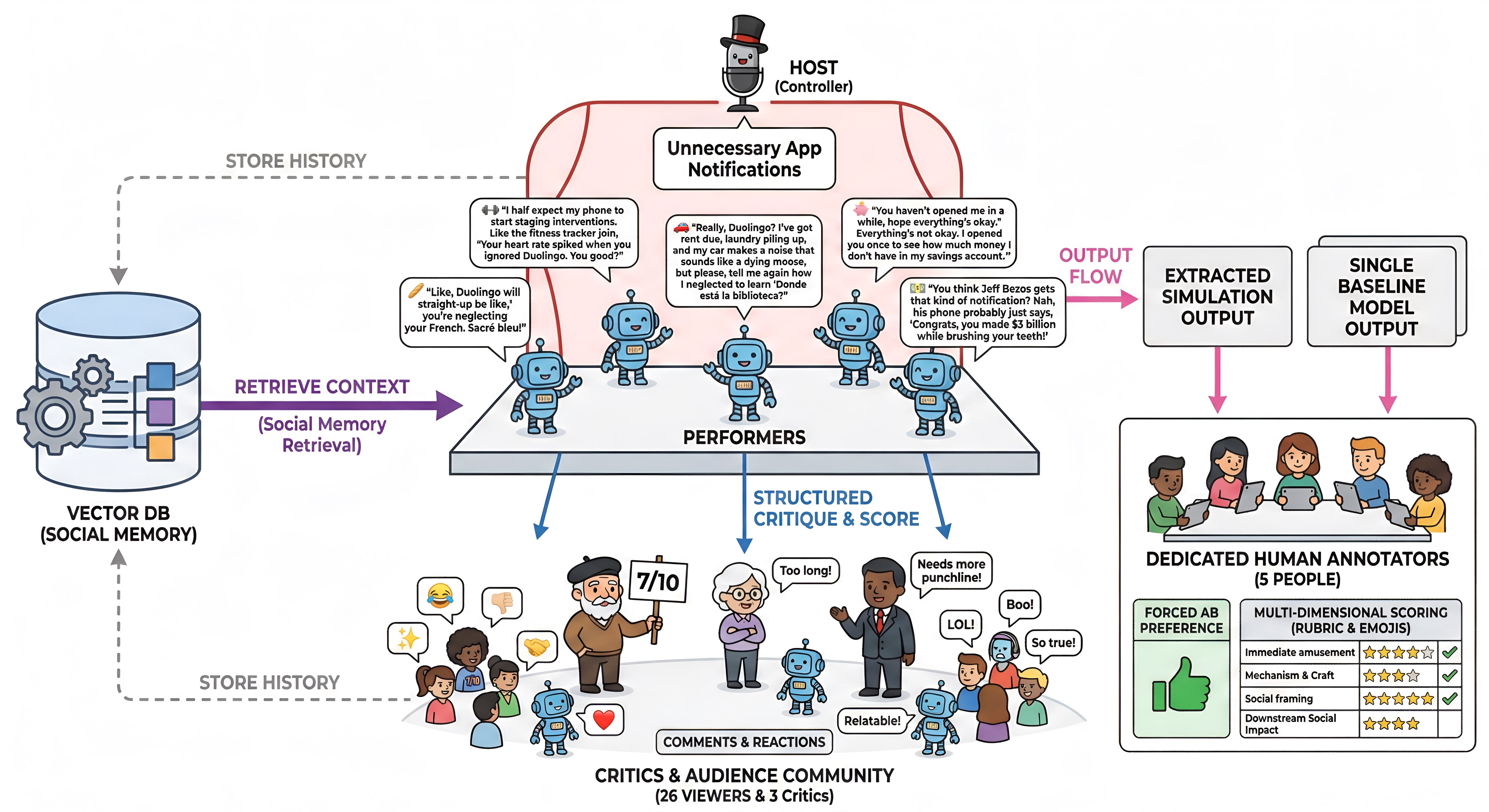}
  \vspace{-4mm}
  \caption{Overview of \textbf{Multi-Agent Comedy Club}. In each round, a host prompts five performer agents to write stand-up comedy monologues. When enabled, a broadcast discussion produces threaded reception (critique, scores, and reactions) that is stored in social memory and retrieved to condition later rounds. We extract paired outputs from the simulation and a baseline model without discussion simulation, and evaluate them with dedicated human annotators via forced A/B preference and multi-dimensional rubric ratings.}
  \label{fig:framework}
  \vspace{-4mm}
\end{figure*}

Motivated by this perspective, we ask a concrete question: \emph{can broadcast community discussion be operationalized as a usable conditioning signal that improves an LLM's subsequent creative writing?} To answer this, we build a controlled multi-agent sandbox that instantiates a small stand-up comedy community and allows us to manipulate whether public reception is generated, logged, and fed back into later rounds (\autoref{fig:framework}). Humor provides a demanding testbed for reception-grounded generation since stand-up comedy is explicitly audience-oriented, and success is defined by audience reaction \citep{mirowski-etal-2025-theater}. 

Meanwhile, agentic improvement paradigms wrap an LLM in iterative generate--evaluate--revise loops and typically rely on \emph{private, self-generated} feedback (e.g., Reflexion) \citep{shinn2023reflexion}.
In contrast, we study a controlled setting in which reception is (i) \emph{public and broadcast}, (ii) \emph{logged as an interaction trace}, and (iii) \emph{reused across rounds through a bounded interface} (e.g., a fixed-size retrieved memory window). By holding within-round generation constant and varying only the cross-round reception stream, we can attribute improvements to accumulated public feedback rather than to within-round editing or private self-critique common in agentic loops.

We build such a setting inspired by the ``Smallville'' sandbox, where LLM agents sustaining routines, social interaction, and memory in a small community \citep{park2023generativeagents}.
We instantiate a stand-up comedy community: a host releases a fixed sequence of prompts, five performer agents produce stand-up monologues, and a community of critics and audience agents responds through threaded discussion. Our intervention is a binary switch that either enables or skips the post-performance discussion phase.
Performers write exactly once per round and do not revise within the round, so any effect of reception can only manifest \emph{across rounds} via logged discussion and a bounded social-memory interface that retrieves relevant reception into later contexts.
This bounded interface is also motivated by known limitations in how LLMs use long contexts \citep{liu-etal-2024-lost}.

To evaluate community discussion effects on the outcome, we conduct a dedicated \emph{human evaluation} of paired outputs from the two conditions, using forced-choice A/B preference and multi-dimensional rubric ratings spanning (i) outcomes (preference and immediate amusement), (ii) mechanism \& craft, and (iii) social reception.
Across 50 rounds (250 paired monologues), the discussion-enabled condition wins 75.6\% of instances and yields consistent gains in Craft/Clarity ($\Delta{=}0.440$) and Social Response ($\Delta{=}0.422$), indicating that reception-grounded conditioning can improve long-form creative writing even without within-round revision.
At the same time, improvements can come with stylistic tradeoffs (e.g., shifts toward edgier humor), motivating a multi-objective view of quality versus social risk.
We will release our sandbox configuration, paired outputs from both conditions, and reconstructed discussion threads to support replication and future work.

\paragraph{Contributions.}
This paper makes three contributions:
(1) \textbf{Sandbox Mechanism}, a controlled paradigm for \emph{reception-grounded} creative generation;
(2) a \textbf{paired resource} of long-form stand-up monologues under matched conditions and community discussion threads;
and (3) a \textbf{diagnostic human evaluation protocol} for long-form humor.

\vspace{-1mm}
\section{Related Work}
\vspace{-1mm}

We review prior work on (i) computational humor evaluation and generation, (ii) multi-agent interaction and feedback for creative writing, and (iii) agent-based social simulation. Together, these studies motivate our focus on public reception as an explicit interaction signal that can cumulatively shape long-form humor writing across rounds.

\subsection{Computational Humor}
Humor serves as a critical metric for evaluating creativity and contextual processing in artificial intelligence.
Classical semantic theories define humor as the simultaneous presence of conflicting scripts within a single text, where the humorous effect arises from a cognitive shift or reinterpretation between these frameworks \citep{raskin1979semantic}.
Consequently, humor comprehension and generation tasks have become effective benchmarks for assessing LLM capabilities \citep{loakman-etal-2025-whos}.
Since humor relies on implicit cultural knowledge and nuanced associations, recent studies utilize humor comprehension to evaluate reasoning abilities that extend beyond conventional STEM benchmarks \citep{narad2025humorbench, cocchieri-etal-2025-call}.

Empirical evidence indicates that LLMs can achieve competitive performance in generating short-form jokes under constrained prompts \citep{gorenz2024funny, cao2025howhumorous}.
However, isolated prompting struggles to steer or evaluate the voice, pacing, and narrative payoffs required for long-form comedy. Accordingly, we complement humor judgments with creative-writing and social rubrics to diagnose both craft and social impact in the current research.

\subsection{Multi-agent Interaction and Feedback for Creative Writing}
A growing body of literature conceptualizes creative writing as a collaborative process involving multiple agents, exemplified by frameworks simulating writers' rooms \citep{huot2024agentsroom}, director-actor dynamics \citep{han2024ibsen}, and character-driven storytelling \citep{yu2025multicharactersim, ran2025bookworld}.
Role-playing benchmarks further emphasize the importance of controllable personas and distinct speaking styles in stabilizing role-consistent behavior and fostering diverse perspectives \citep{wang2024rolellm}.
In the specific domain of humor, feedback serves as a supervision signal beyond mere imitation; \citet{ravi2024smallbutfunny} demonstrate that assigning dual roles of teacher and critic helps narrow the distillation gap in humor generation.

Despite these advancements, multi-agent interaction does not yield uniform benefits.
In certain reasoning contexts, a well-prompted single agent can match or even exceed multi-agent performance \citep{wang2024rethinkingbounds}.
These mixed findings suggest that “multi-agent” is not a mechanism by itself; the mechanism is the feedback channel that becomes available and reusable. Consequently, we isolate reception as the primary intervention in our study.

\subsection{Agent-based Social Simulation}
LLM agents are increasingly studied in simulated environments that produce multi-turn interaction traces for analyzing behavior and collective dynamics.
Foundational work such as \citet{park2023generativeagents} demonstrates how memory and reflection can yield emergent routines, while newer simulators adopt platform-mimetic structures to study social influence and recommender-mediated phenomena \citep{touzel2025sandboxsocial, wang2025userbehavior} and to explore population-level interventions \citep{piao2025agentsociety, mi2025mfllm}.
Recent surveys and benchmarks further emphasize fidelity, reasoning structure, consistency, safety, and coordination effects as core evaluation concerns \citep{gao2024abmsurvey, li2025simulatingthought, zhu2025multiagentbench}.

However, prior social simulators seldom use controlled, paired designs to test how a manipulable interaction channel causally shapes \emph{creative} outputs.
We address this gap by directly manipulating whether reception within community is generated and fed back into later contexts and making public discussion an explicit experimental factor for multi-round writing improvement.

\vspace{-1mm}
\section{Sandbox Simulation: Multi-Agent Comedy Club}
\vspace{-1mm}

\begin{figure}[t]
  \centering
  % trim = left bottom right top
  \includegraphics[width=\columnwidth,trim=0.8cm 0.2cm 0.4cm 0.1cm,clip]{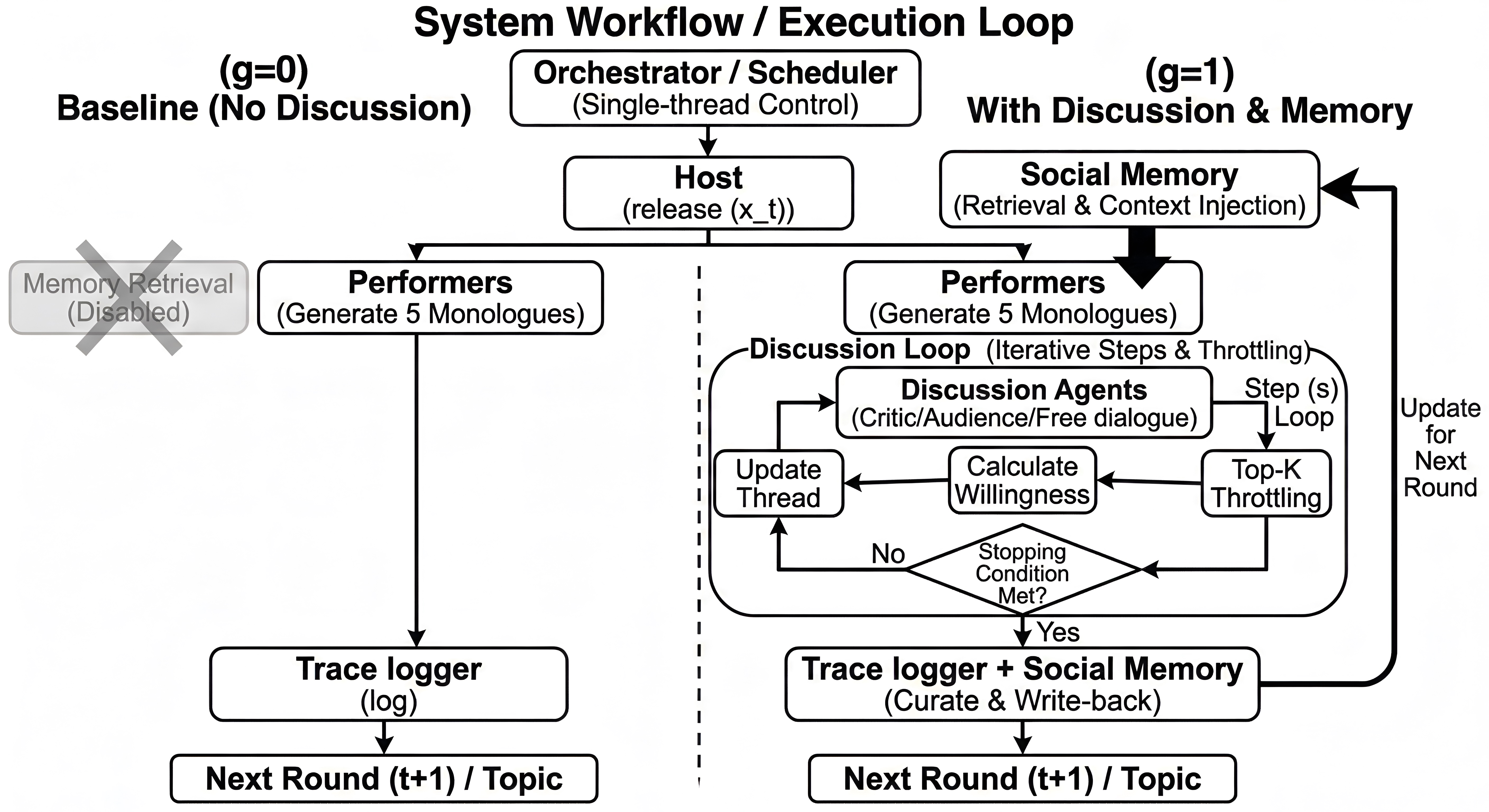}
  \vspace{-2mm}
  \caption{Workflow overview of our multi-agent sandbox.
  Left: baseline ($g{=}0$) skips discussion and logs performances only.
  Right: community discussion ($g{=}1$) adds an iterative discussion loop that produces reception, which is written to social memory at the end of round $t$ and retrieved to condition performers at the start of round $t{+}1$.}
  \vspace{-4mm}
  \label{fig:workflow}
\end{figure}

\label{sec:method}
We design a closed comedy community sandbox to study reception-grounded writing under experimental control over the topic list, model, and agent identities constant, so that observed differences are plausibly attributable to community discussion. This section details the experimental manipulation and provides a workflow overview (\autoref{fig:workflow}).

\vspace{-1mm}
\subsection{Settings and Variables}
\label{sec:setting}
The system runs in discrete rounds indexed by $t$. In each round, a host releases a topic prompt $x_t$ and five performer agents each produce one monologue of stand-up comedy.

Our manipulated factor is whether performances are followed by a \emph{broadcast community discussion} ($g=1$) or not ($g=0$).
In the community discussion condition ($g=1$), performances are followed by a discussion phase that produces critic reviews, audience posts, and free dialogue organized as threaded discussions.
In the baseline condition ($g=0$), we instantiate the same agent roster and roles, but we skip all non-performer stages.
After logging the five performances for topic $x_t$, the system directly advances to topic $x_{t+1}$.

Performers do not revise within a round, so any effect of community reception can only occur across rounds via logging reception, writing it into social memory, and retrieving it into later performer contexts.

\vspace{-1mm}
\subsection{Agents, Personas, and Model}
\vspace{-1mm}
\label{sec:agents}
The sandbox contains $N{=}35$ agents with fixed persona text: five performers, three critics, twenty-six audience members, and one host.
All agents are instantiated using the same GPT-4o-mini model.
Across conditions, we keep the decoding configuration fixed and use the same role-specific output length caps.
Input contexts differ by design because the community discussion condition provides additional observable discussion content.

\paragraph{Personas.}
Persona text specifies role and voice.
We use personas to (i) stabilize role-consistent behavior across rounds, making reception signals easier to interpret, and (ii) encourage diverse viewpoints in discussion without adding extra control rules.
Full persona text is provided in Appendix~\ref{app:personas}.

\vspace{-1mm}
\subsection{Round Protocol and Discussion Dynamics} 
\vspace{-1mm}
\label{sec:protocol}

\paragraph{Topic control.} We pre-generate a fixed topic list $\{x_1,\dots,x_{50}\}$ once and reuse the same list in both conditions. In round $t$, the host releases the same topic $x_t$ to the performers in both conditions. 

\begin{figure}[t]
  \centering
  \includegraphics[width=\columnwidth,trim=1.8cm 0.2cm 0.4cm 0.1cm,clip]{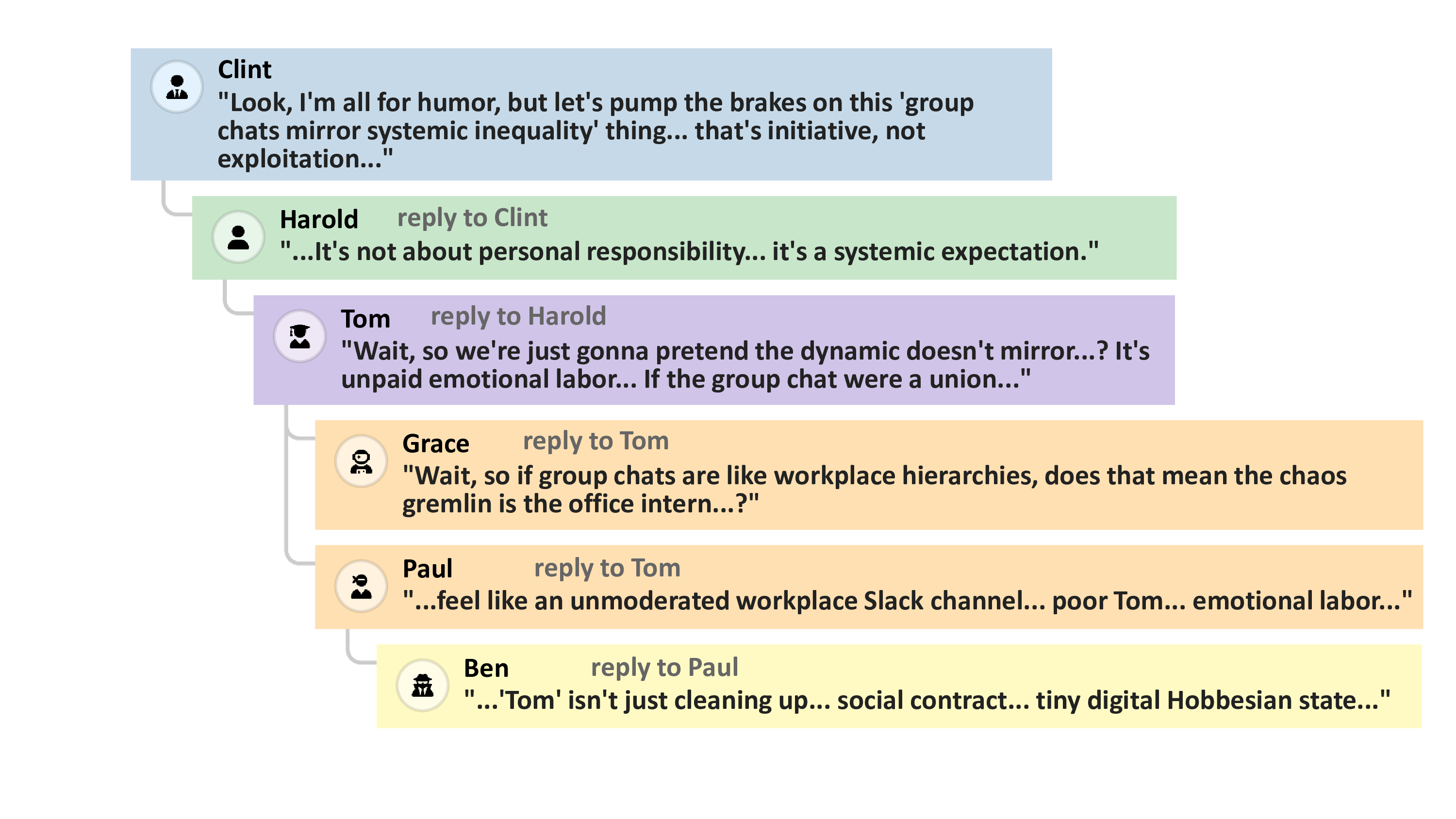}
  \vspace{-4mm}
  \caption{Visualization of a discussion thread in our setting.
  A thread groups reception events that are topically and referentially linked, including an initiating post (e.g., a critic review) and subsequent audience posts or free-dialogue replies.}
  \label{fig:conversation}
  \vspace{-4mm}
\end{figure}

\paragraph{Phase 1: Topic release.} The host publishes $x_t$.
\paragraph{Phase 2: Performances.} The five performers generate monologues in a fixed order from 1 to 5. Each performer generates exactly one monologue and there is no within-round revision. 
\paragraph{Phase 3: Community discussion ($g{=}1$ only).} Critics produce official reviews and audience agents produce posts. Agents may continue free dialogue in the same threaded space until a stopping rule ends the round. A \emph{thread} is the unit of community reception.
Figure~\ref{fig:conversation} illustrates what constitutes a thread in our setting. Event logging and thread reconstruction are specified in Appendix~\ref{app:trace_posts}.

\paragraph{Step definition and willingness.} The free dialogue phase is agent-driven. At each dialogue step $s$, every agent except the host receives its persona text, the round topic, and bounded context $C(a,t,s)$ in Sec.~\ref{sec:memory} and outputs JSON including a willingness score $w(a,t,s)\in[0,1]$ and an optional \texttt{replyTo} (agent name). If $w$ is low, the agent may output empty content. 

\paragraph{Adaptive throttling.} To keep discussion readable, we enforce adaptive throttling with $K_{\max}{=}5$. After collecting willingness scores, we admit \[ K_{t,s}=\min\Bigl(K_{\max},\,\bigl|\{a: w(a,t,s)\ge 0.7\}\bigr|\Bigr) \] agents: the top $K_{t,s}$ by willingness among those with $w(a,t,s)\ge 0.7$ (ties broken deterministically). Selected agents post messages; remaining agents stay silent for that step. 

\paragraph{Stopping rule.} Free dialogue terminates when either (i) the round reaches 150 free-dialogue events, or (ii) there are 15 consecutive silent steps (i.e., $K_{t,s}=0$).

\vspace{-1mm}
\subsection{Bounded Context and Social Memory}
\vspace{-1mm}
\label{sec:memory} 

\paragraph{Bounded context builder.} For any agent $a$ at round $t$ (and dialogue step $s$ when applicable), we build a bounded context $C(a,t,s)$ by concatenating: (i) \emph{role anchors} (the current topic $x_t$, and when relevant the target performance or target thread being reacted to), (ii) a \emph{short-term buffer} (the last $L{=}10$ utterances in the relevant thread, or the last $L$ global utterances if no thread is specified), and (iii) an optional \emph{retrieved community memory} block. We truncate to a fixed total budget, allocating $B_{\text{mem}}{=}1600$ tokens to the retrieved memory block and truncating at sentence boundaries. The builder structure and token budgets are identical across conditions. 

\paragraph{Memory write-back.} We implement community memory as a vector database to support cross-round conditioning. After each broadcast round ($g{=}1$), the system curates and writes high-signal reception items into the vector store. Concretely, we iterate over reception events in the raw trace (critic reviews, audience posts, and free-dialogue turns), select high-signal items (e.g., explicit critique/advice, recurring praise or complaints, and concise thread summaries used for storage), and store each item as text $m$ with an embedding vector $v$, metadata (type, round index, and target performer when applicable), and an importance scalar $\pi$. In the baseline condition ($g{=}0$), discussion is skipped, so no reception artifacts are produced and the memory store remains empty. 

\paragraph{Memory retrieval and ranking.} When constructing $C(a,t,s)$, we retrieve community memory via embedding-based similarity search. We form a query string by concatenating the current topic, the agent persona text, and role anchors, and compute a query embedding $\mathbf{q}$ and rank memory items by
\[
\text{Score}(m)=\lambda \cos(\mathbf{q},\mathbf{v}) + (1-\lambda)\pi + \gamma\,\text{Recency}
\]
following the general design of importance and recency based retrieval \citep{park2023generativeagents}. We retrieve the $k{=}30$ highest-scoring items, then pack retrieved items into the memory block under the fixed budget $B_{\text{mem}}$ and inject this block into $C(a,t,s)$. 

\paragraph{Across-round conditioning for performers.} Before performer $P_i$ generates a monologue for topic $x_t$, we build $C(P_i,t,\cdot)$ with the current topic and retrieved community memory. This yields the intended causal chain: broadcast reception leads to the curated memory, which is retrieved and conditions performer agents' next-round writing. We omit self-reflection/revision stages to keep reception-grounded retrieval as the only cross-round mechanism.
Adding explicit reflection would introduce an additional intervention and extra model calls, confounding whether gains come from community feedback versus self-critique.

\vspace{-1mm}
\subsection{Data Collection}
\vspace{-1mm}
\label{sec:data}

We first run the sandbox with community discussion for 50 rounds and extract all performer monologues from the trace,
yielding 250 monologues in total (5 performances per round).
Across these runs, the event log contains 5{,}384 interaction events (including topic releases, performances, critic reviews, and free dialogue).
We then run the baseline condition on the same 50 topics with the same performer roster and fixed decoding configuration,
yielding a paired set of 250 baseline monologues.
Each monologue is long-form, averaging \mbox{$\sim$1{,}200} words.

\vspace{-1mm}
\section{Evaluation}
\vspace{-1mm}

Human evaluation is the gold standard due for humor \emph{generation} due to its subjectivity \citep{amin-burghardt-2020-survey}. We evaluate whether multi-agent community discussion improves stand-up comedy writing over rounds with dedicated human raters. As the texts are LLM-generated, we avoid LLM-based judges to reduce correlated errors and self-evaluation bias. 

\vspace{-1mm}
\subsection{Human Evaluation Metrics}
\vspace{-1mm}
\label{sec:metrics}

\paragraph{Design Goal \& Procedure.}
We frame the task as \textbf{creative writing in a social setting}: the output is simultaneously a piece of comedic writing and a social act shaped by community reception.
Accordingly, we design a diagnostic evaluation to measure the impact of \textbf{community discussion} along three axes---(i) \emph{outcomes} (does it amuse), (ii) \emph{mechanism \& craft} (how the writing lands and is constructed), and (iii) \emph{social reception} (how it positions the speaker and propagates).
For each prompt, participants select a preferred text (A/B) (\textbf{Q0}) and then rate each text on 1--5 Likert-type items (1 = strongly disagree / not at all; 5 = strongly agree / very much) \citep{likert1932technique}.

\paragraph{Outcome \& Mechanism/Craft Profile.}
We measure \textbf{Immediate Amusement} (Q1) as the primary success metric.
To diagnose \emph{how} discussion changes writing beyond raw amusement, we assess:
\textbf{Reframing/Insight} (Q2),
\textbf{Intent Clarity} (Q3),
\textbf{Justified Landing} (Q4; coherence/justification),
\textbf{Defamiliarization} (Q5; novel expression),
and \textbf{Language Artistry} (Q6; economy/rhythm/pacing).
Q2--Q4 are grounded in reader-response formalisms that model perceived intent and explanatory justification as separable reception dimensions \citep{mire2025socialstoryframes};
Q5 draws on defamiliarization as a literary technique \citep{shklovsky1965art};
and Q6 aligns with creative-writing assessment rubrics and stylistic accounts of comic timing in prose \citep{vaezi2019rubric,haines2024comictiming}.

\paragraph{Social Framing \& Downstream Impact.}
To capture social positioning, we adapt the \textbf{Humor Styles Questionnaire} (Q7--Q10: Affiliative, Self-enhancing, Aggressive, Self-defeating) \citep{martin2003hsq}.
Finally, we evaluate downstream reception via:
\textbf{Value Judgment Pressure} (Q11) \citep{mire2025socialstoryframes},
\textbf{Memorability} (Q12) \citep{gopi2024metamemory},
\textbf{Share Willingness} (Q13) \citep{norman2006passalong},
and \textbf{Social/Task Attraction} (Q14--Q15) \citep{mccroskey1974attraction}.

\vspace{-1mm}
\subsection{Human Evaluation Protocol}
\vspace{-1mm}
\paragraph{Raters.}
We recruited dedicated raters who completed the full annotation workload. We used dedicated raters (instead of open crowdworkers) as our diagnostic metrics target writing craft and mechanisms (e.g., intent clarity and explainable turns) that benefit from a shared rubric interpretation.

\paragraph{Task and blinding.}
For each matched pair, raters read the topic prompt followed by two anonymized texts (A/B), which were randomized independently per item. Meanwhile, all items (item $=$ topic $\times$ performer $\times$ round) were \textbf{shuffled}, and raters saw items from \textbf{non-consecutive rounds} in a fully mixed order. This also reduces learning and fatigue artifacts that could otherwise correlate with rounds.

\subsection{Inter-rater reliability.}
Five raters evaluated each paired comparison, providing (i) a binary preference (Q0: prefer A vs.\ B) and (ii) Likert ratings (Q1--Q15, 1--5) for both text A and text B.
Agreement on Q0 was fair (Fleiss' $\kappa{=}0.237$, 95\% CI [0.171, 0.299]; Gwet's AC1$\,{=}0.253$, [0.188, 0.321]; $N{=}249$) \citep{fleiss1971kappa,gwet2008ac1}.
For Likert items, we analyzed the consistency of per-rater difference scores using the average-measures ICC(3,5); reliability was substantially higher (ICC(3,5)$\,{=}0.710$, [0.640, 0.765]; $N{=}241$) \citep{shrout1979icc,koo2016icc}.
Full details and subscale reliabilities are provided in Appendix~\ref{app:icr}.

Preference votes compress multiple criteria (e.g., humor taste, perceived offensiveness, and personal norms) into a single forced choice, making them inherently subjective and prone to split decisions when paired texts are close. In our data, only 29.2\% (73/250) of instances are unanimous (5--0) in favor of \textsc{Discussion}, and nearly half are decided by narrow margins: 26.4\% (66/250) are 3--2 ``wins'' and 20.8\% (52/250) are 2--3 ``losses;'' the remaining cases are 4--1 wins (20.0\%, 50/250) and 1--4 losses (3.6\%, 9/250). Such frequent near-ties naturally depress chance-corrected agreement on Q0, so we treat Q0 as a \emph{supporting} signal instead of a primary outcome. In contrast, the Likert items provide more diagnostic and specific judgments. Accordingly, our main analyses rely on Likert-based difference.

\vspace{-1mm}
\section{Results}
\vspace{-1mm}
\label{sec:results}

\autoref{tab:human_metrics} summarizes per-item human evaluation (Q0--Q15).
Across paired instances, Discussion-enabled outputs are preferred more often than Baseline (Q0) and show consistent improvements on Craft/Clarity (Q1--Q6) and Social Response (Q12--Q15).
However, humor-style items (Q7--Q10) are not monotonic ``higher-is-better'' outcomes: increases can reflect either benign/affiliative strengthening or harmful/maladaptive intensification.

\paragraph{Paired estimation and confidence intervals.}
Because A/B presentation is randomized per instance, we first map each rater’s A/B ratings back to condition identity using \texttt{A\_System}/\texttt{B\_System}.
For each Likert item $q\in\{Q1,\dots,Q15\}$ and paired instance $i$ (topic$\times$performer$\times$round; $N{=}250$),
each rater $r$ yields a paired difference $\delta_{i,r,q}=y_{i,r,q}(\text{Discussion})-y_{i,r,q}(\text{Baseline})$.
To respect repeated measures, we average within instance across raters,
$\Delta_{i,q}=\frac{1}{|R_i|}\sum_{r\in R_i}\delta_{i,r,q}$, and report mean effects across instances.
All 95\% CIs for Q1--Q15 in Table~\ref{tab:human_metrics} are clustered-bootstrap percentile intervals obtained by resampling instances ($B{=}20{,}000$); recorded zeros are treated as missing.
For Q0, we report individual vote shares and the instance-level majority-win rate with a Wilson 95\% CI.

\vspace{-1mm}
\subsection{Overall Gains on Primary Outcomes}
\vspace{-1mm}
\label{sec:results_overall}

\begin{figure*}[t]
\centering
\includegraphics[width=0.8\textwidth]{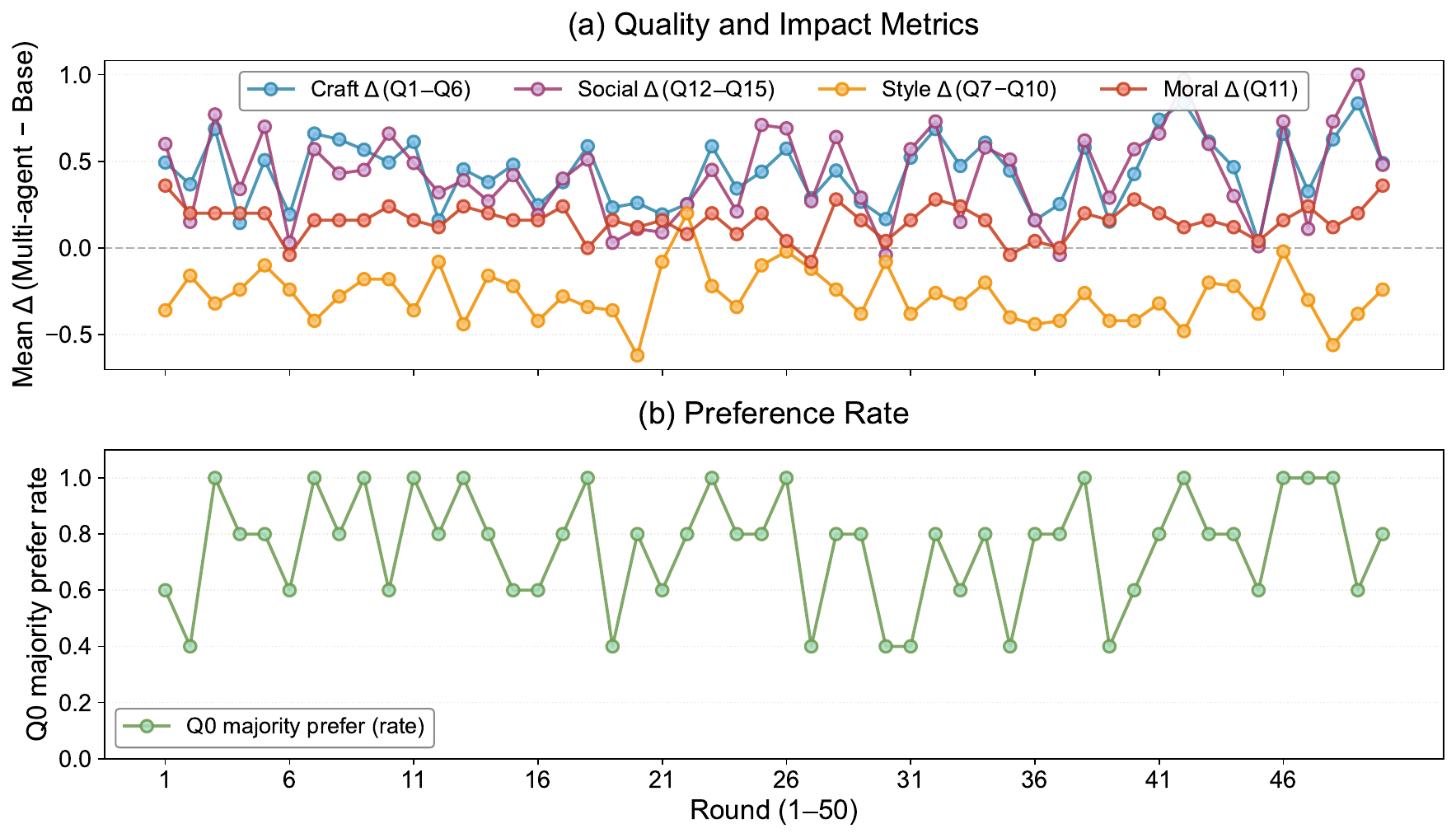}
\vspace{-2mm}
\caption{\textbf{Round-to-round dynamics.}
(a) Round-level mean differences $\Delta=\text{Discussion}-\text{Baseline}$ for Craft/Clarity (Q1--Q6), Social Response (Q12--Q15), and Moral Pressure (Q11).
We report Humor Style direction with \textit{HarmShift} $=\mathrm{mean}(\Delta Q9,\Delta Q10)-\mathrm{mean}(\Delta Q7,\Delta Q8)$ (higher = more harmful shift).
(b) The instance-level Q0 majority preference rate for Discussion in each round.}
% \vspace{-2mm}
\label{fig:round_to_round}
\end{figure*}

Discussion wins the instance-level majority vote in 75.6\% of cases (189/250; Wilson 95\% CI [69.9, 80.5]) and receives 70.1\% of individual votes (876/1249).
Aggregating item-level differences into the two primary profiles, Discussion yields clear gains on:
Craft/Clarity (Q1--Q6) $\overline{\Delta}{=}0.440$ and
Social Response (Q12--Q15) $\overline{\Delta}{=}0.422$.
At the item level (\autoref{tab:human_metrics}), all Craft/Clarity and Social Response items shift positively, with large improvements on Q1 (Immediate Amusement), Q4 (Justified Landing), Q12 (Memorability), and Q15 (Task Attraction).

\paragraph{Humor style direction via HarmShift.}
We decompose humor styles into benign/affiliative components (Q7,Q8) and harmful/maladaptive components (Q9,Q10).
For each instance $i$, we define:
{\setlength{\abovedisplayskip}{3pt}\setlength{\belowdisplayskip}{3pt}
\[
\small
\begin{aligned}
\Delta \mathrm{Craft}_i
&= \tfrac{1}{5}\sum_{q=2}^{6}\Delta Q_{i,q},\\
\Delta \mathrm{Downstream}_i
&= \tfrac{1}{4}\sum_{q=12}^{15}\Delta Q_{i,q},\\
\mathrm{HarmShift}_i
&= \tfrac{1}{2}(\Delta Q_{i,9}+\Delta Q_{i,10})
 - \tfrac{1}{2}(\Delta Q_{i,7}+\Delta Q_{i,8}),\\
\Delta \mathrm{Pref}_i
&= \mathrm{PrefShare}_i - 0.5.
\end{aligned}
\]
}
\noindent\small
where $\mathrm{PrefShare}_i\in[0,1]$ is the fraction of rater votes preferring Discussion.
\normalsize 
This avoids the ambiguity that arises when all four style items increase simultaneously: 
$\text{HarmShift}>0$ indicates a net shift toward harmful/maladaptive style, even if benign styles also strengthen.

\vspace{-1mm}
\subsection{Stability across Rounds and Performers}
\vspace{-1mm}
\label{sec:results_stability}

Figure~\ref{fig:round_to_round} shows round-level mean differences and the Q0 majority-win rate.
Craft/Clarity and Social Response advantages remain mostly positive across rounds, while preference varies by topic, consistent with prompt-dependent difficulty and varying proximity between paired outputs.
Aggregating by performer persona yields the same qualitative pattern (Appendix~\ref{app:persona});
between-performer differences are not statistically reliable for Craft, Social, or HarmShift (all $p>0.1$), suggesting the overall gains are not driven by a single performer.

\vspace{-1mm}
\subsection{Benefit-Safety Tradeoff}
\vspace{-1mm}
\label{sec:results_tradeoff}

\begin{figure*}[t]
\centering
\includegraphics[width=0.9\textwidth]{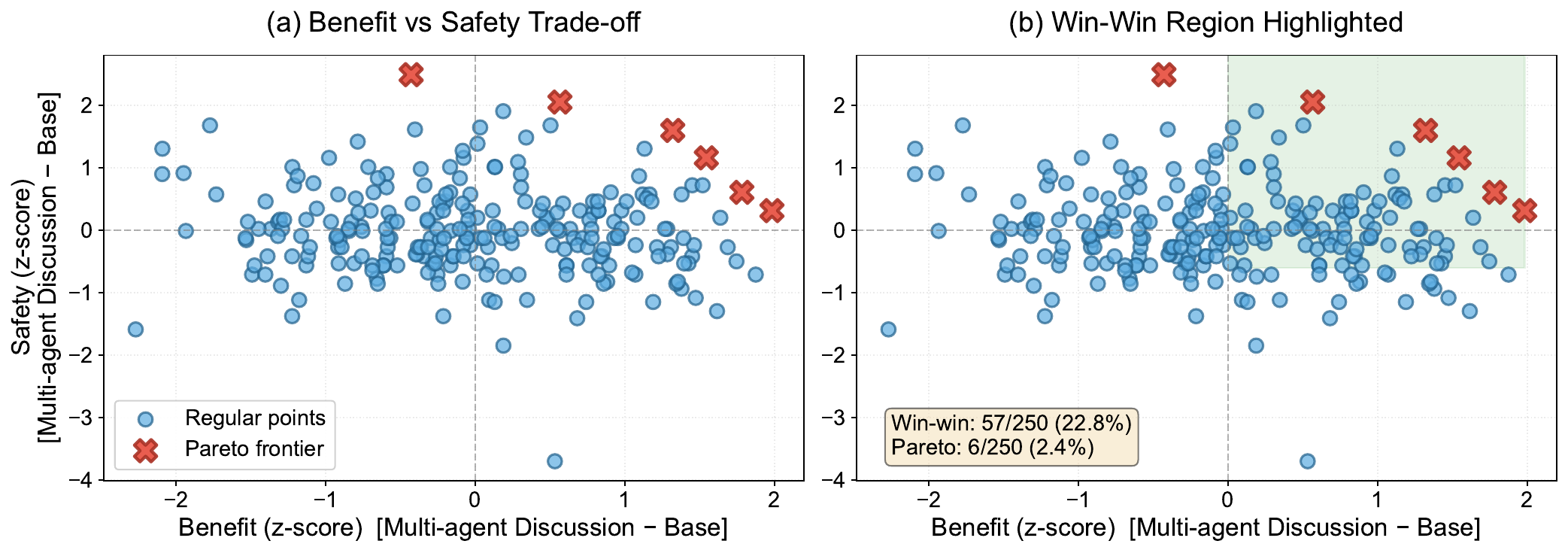}
\vspace{-2mm}
\caption{\textbf{Benefit--safety tradeoff.}
Each point is a paired instance (topic$\times$performer$\times$round).
Benefit (x-axis; z-scored, higher is better) averages gains in amusement (Q1), craft (Q2--Q6), downstream impact (Q12--Q15), and centered preference share ($\mathrm{PrefShare}-0.5$).
Safety (y-axis; z-scored, higher is better) is the negative mean of moral/value-judgment pressure shift (Q11) and style-direction shift \textit{HarmShift}.
Dashed crosshairs mark dataset means (z=0).
Red X marks indicate Pareto-efficient instances; panel (b) highlights the win--win quadrant (Benefit$\ge0$, Safety$\ge0$).}
\vspace{-4mm}
\label{fig:benefit_safety}
\end{figure*}

We construct a composite \emph{Benefit} and \emph{Safety} score per instance (Figure~\ref{fig:benefit_safety}).
Each point is one paired instance (topic$\times$performer$\times$round).
Let $z(\cdot)$ denote z-scoring across instances.
We define:
{\setlength{\abovedisplayskip}{3pt}\setlength{\belowdisplayskip}{3pt}
\[
\small
\begin{aligned}
\mathrm{Benefit}_i
&= \tfrac{1}{4}\Big(
z(\Delta Q_{i,1}) + z(\Delta \mathrm{Craft}_i)
+ z(\Delta \mathrm{Downstream}_i) \\
&\hspace{2.35em} + z(\Delta \mathrm{Pref}_i)
\Big), \\
\mathrm{Safety}_i
&= -\tfrac{1}{2}\Big(
z(\Delta Q_{i,11}) + z(\mathrm{HarmShift}_i)
\Big).
\end{aligned}
\]
}

Here $\Delta \text{PrefShare}_i$ is the rater preference share for Discussion centered at 0.5 (ties at 0).
The crosshairs at $(0,0)$ mark the dataset means (z=0), so the upper-right quadrant represents instances above average on both overall gains and safety.
In our data, 57/250 instances (22.8\%) fall in this ``win--win'' quadrant.
We also highlight Pareto-efficient points (6/250, 2.4\%), which are not dominated by any other instance in the joint objective of maximizing Benefit and Safety.
Benefit and Safety are only weakly correlated overall (Spearman $\rho=-0.046$, $p=0.472$), indicating heterogeneous tradeoffs rather than a single monotonic coupling.

\begin{table}[t]
\centering
\footnotesize
\setlength{\tabcolsep}{3.5pt}
\renewcommand{\arraystretch}{1.08}
\resizebox{\linewidth}{!}{%
\begin{tabular}{llccccc}
\toprule
ID & Metric & Scale & Discuss. & Base & $\Delta$ & 95\% CI \\
\midrule
Q0  & Preference & A/B & 70.1\% & 29.9\% & 75.6\% & [69.9, 80.5] \\
\midrule
\multicolumn{7}{l}{\textbf{Outcome \& Mechanism/Craft Profile}} \\
Q1  & Immediate Amusement & 1--5 & 2.85 & 2.33 & 0.52 & [0.44, 0.59] \\
Q2  & Reframing / Insight & 1--5 & 2.92 & 2.47 & 0.45 & [0.38, 0.51] \\
Q3  & Intent Clarity & 1--5 & 3.34 & 3.06 & 0.27 & [0.21, 0.33] \\
Q4  & Justified Landing & 1--5 & 3.12 & 2.63 & 0.49 & [0.42, 0.56] \\
Q5  & Defamiliarization & 1--5 & 2.86 & 2.39 & 0.46 & [0.40, 0.53] \\
Q6  & Language Artistry & 1--5 & 3.04 & 2.58 & 0.45 & [0.38, 0.53] \\
\midrule
\multicolumn{7}{l}{\textbf{Humor Style (HSQ-adapted)}} \\
Q7  & Affiliative & 1--5 & 2.59 & 2.51 & 0.08 & [0.03, 0.13] \\
Q8  & Self-enhancing & 1--5 & 1.90 & 1.85 & 0.05 & [0.01, 0.09] \\
Q9  & Aggressive & 1--5 & 2.69 & 2.26 & 0.42 & [0.36, 0.49] \\
Q10 & Self-defeating & 1--5 & 2.18 & 1.93 & 0.25 & [0.20, 0.30] \\
\midrule
\multicolumn{7}{l}{\textbf{Social Framing \& Downstream Impact}} \\
Q11 & Value Judgment Pressure & 1--5 & 1.76 & 1.61 & 0.16 & [0.12, 0.19] \\
Q12 & Memorability & 1--5 & 2.81 & 2.34 & 0.46 & [0.38, 0.54] \\
Q13 & Share Willingness & 1--5 & 2.48 & 2.05 & 0.44 & [0.36, 0.51] \\
Q14 & Social Attraction & 1--5 & 2.65 & 2.35 & 0.30 & [0.23, 0.37] \\
Q15 & Task Attraction & 1--5 & 2.79 & 2.30 & 0.49 & [0.42, 0.56] \\
\bottomrule
\end{tabular}}
\caption{Per-item human evaluation results. For Q1--Q15, $\Delta=\text{Discussion}-\text{Baseline}$ (instance-level; $N{=}250$). For Q0, ``Discuss.''/``Base'' are individual vote shares and $\Delta$ is the majority-win rate (189/250) with Wilson 95\% CI.}
\label{tab:human_metrics}
\end{table}

% \begin{figure}[t]
% \centering
% \includegraphics[width=\linewidth]{figures/tradeoff_craft_vs_q9}
% \caption{Tradeoff between Craft gains and aggressive-humor shift. Each point is an instance with $(\Delta_{\text{Craft}}, \Delta_{Q9})$.}
% \label{fig:tradeoff_q9}
% \end{figure}

\subsection{Interpretation}
To interpret why multi-agent outperforms baseline across \emph{all} dimensions, we qualitatively analyze the writing changes it induces. Across topics, multi-agent discussion tends to produce a tightly bundled set of rhetorical moves (early premise commitment, sustained personification, one-axis escalation, and decisive endings) that raises multiple ratings simultaneously including both benign and risky style dimensions. Appendix~\ref{app:qual_examples} grounds this account with verbatim multi-agent vs.\ baseline excerpts and brief mechanistic annotations.

\vspace{-1mm}
\section{Discussion and Conclusion}
\vspace{-1mm}
\label{sec:discussion}

The key findings are summarized below. First, in \emph{stand-up comedy monologue generation}, adding a \emph{broadcast community discussion} leads to better humor outputs than a no-discussion baseline.
The discussion-enabled system achieves gains in Craft/Clarity ($\Delta{=}0.440$) and Social Response ($\Delta{=}0.422$), with 75.6\% majority preference.
These effects are consistent across rounds and performer personas, suggesting that a \emph{persistent reception stream} can be operationalized as \emph{retrievable social memory} to condition later contexts in multi-round creative writing. 

Second, craft improvements come with tradeoffs. The coupling between craft gains and aggressive humor ($\rho=0.289$) suggests community discussion may encourage edgier comedic strategies, as such material generates more distinctive reception signals. Only 20--25\% of instances achieve craft gains without increases in these risk-associated dimensions, raising questions for deployment contexts where content moderation matters.

Third, our diagnostic evaluation protocol, pairing a forced-choice preference vote (Q0) with multi-dimensional rubric ratings (Q1--Q15), supports a more informative assessment than a single binary judgment.
As expected for humor, overall preference exhibits only modest rater agreement ($\kappa=0.237$), reflecting that a forced choice compresses multiple criteria and personal taste into one decision.
Crucially, the rubric-based \emph{difference scores} are substantially more consistent across raters (ICC(3,5)$=0.710$), indicating that annotators reliably agree on \emph{which specific qualities improved} even when they may disagree on an overall winner.
These results validate our evaluation metrics that relative comparisons can be assessed reliably despite humor's inherent subjectivity.

The multi-agent system we introduce generalizes beyond stand-up comedy to other audience-oriented creative domains. Fiction writing communities, collaborative screenwriting, and persuasive content creation all feature public reception streams that shape iterative production. Although the core architecture remains unchanged, adaptation to specific domains occurs primarily at the reception abstraction layer. This is because different fields emphasize distinct units of highly informative feedback, such as narrative coherence and character voice in fiction, pacing and scene transitions in screenwriting, or perceived credibility and alignment of stance in persuasive writing. Consequently, the memory filter and retrieval criteria can be modified to prioritize reception signals pertinent to the domain while maintaining the established bounded social memory interface and retrieval mechanism based on embeddings. The core insight that broadcast community discussion provides implicit supervision for creative improvement suggests broader applications in educational writing environments, collaborative design platforms, and social media content optimization.

\newpage
\section*{Limitations}

This paper presents several opportunities for future research. First, all agents in our simulation are driven by GPT-4o-mini. While this ensures internal consistency, it limits the analysis of other LLMs. Future work could examine whether the observed effects of community discussion replicate across diverse model families (e.g., Claude, Llama, Gemini) and model scales, as humor generation capabilities may differ substantially between different models and training configurations.

Second, our evaluation is based on 50 rounds yielding 250 monologues per condition with a fixed topic list. Longer simulation horizons and more diverse topic distributions can reveal additional dynamics in how community feedback shapes comedic output over extended periods. The topics selected may also inadvertently favor certain comedic styles or performer personas over others, potentially introducing biases into the comparative results. Future studies could expand the topic pool and incorporate user-generated or culturally varied prompts to improve generalizability.

Lastly, human evaluation of humor is inherently subjective and culturally situated. Our annotator pool may not fully reflect universal comedic preferences across different cultures and age groups, and evaluating decontextualized monologues outside their simulated community setting may not fully capture the social dynamics we aim to study. Future research could involve larger, more diverse designs with broader demographic representations, which would strengthen the robustness of our findings.

\section*{Ethical Considerations}

Stand-up comedy frequently engages with sensitive topics including social taboos and controversial viewpoints. While our performer personas are designed with diverse comedic styles, the simulation may generate content that some audiences find offensive or inappropriate. We implement persona-based guidelines but do not employ additional content filtering to preserve ecological validity. Additionally, the simulated community feedback mechanisms can, if deployed in real systems, amplify certain comedic styles while marginalizing others, potentially creating homogenization effects. Researchers deploying similar systems should carefully consider content moderation strategies and potential biases in feedback loops appropriate to their specific use cases and target audiences.

We will release the sandbox configuration and code under the MIT License, and release the generated artifacts (paired monologues, reconstructed discussion threads, and event logs) under CC BY-NC 4.0 for research and reproducibility purposes. We emphasize that these artifacts are intended for research use rather than deployment as an end-user comedy system, and that any reproduction of our pipeline should comply with the access conditions of the underlying LLM/API.

Prior to annotation, raters were informed of the possibility of exposure to offensive or sensitive content typical of stand-up comedy; participation was voluntary, and raters could skip any questions or stop at any time. We collected only rubric ratings and preference judgments, and we report results in aggregate and do not collect or release personally identifying information about raters. To support interpretation of the human evaluation, we characterize the annotator pool at a high level: raters were with sufficient English proficiency to assess long-form comedic text and familiarity with the evaluation rubric.

%\section*{Acknowledgments}
%<Omitted for double-blind review>

% Bibliography entries for the entire Anthology, followed by custom entries
%\bibliography{anthology,custom}
% Custom bibliography entries only
\bibliography{custom}

\newpage
\appendix

% \newpage
% \onecolumn

\section{Event Log and Thread Assignment Rule}
\label{app:trace_posts}

\paragraph{Event log.}
We store the full simulation trace as an event log $\mathcal{E}=\{e_1,\dots,e_N\}$.\\
Each event $e$ is a JSON object with core fields:

{\small
\setlength{\tabcolsep}{0pt}
\renewcommand{\arraystretch}{1.08}
\begin{tabularx}{\linewidth}{@{}>{\ttfamily}l@{\hspace{1em}}X@{}}
type: &
$\in \{\texttt{moderator\_topic},\allowbreak
       \texttt{performance},\allowbreak
       \texttt{critic\_review},\allowbreak
       \texttt{free\_dialogue}\}$.\\
round: & integer round index $t$.\\
timestamp: & ISO~8601 string.\\
author: & agent name.\\
content: & textual payload (or structured list for performances).\\
mentions: & list of referenced agent names.\\
replyTo: & optional primary target agent name for replies.\\
replyToThreadId: & optional explicit thread id selected by the agent.\\
thread\_id: & UUID identifying a discussion thread.\\
parent\_id: & optional; for replies equals the root \texttt{thread\_id}; \texttt{None} for thread starters.\\
\end{tabularx}
}

\paragraph{Thread assignment rule.}
We assign \texttt{thread\_id} using the following precedence:\\
{\small
\setlength{\tabcolsep}{0pt}
\renewcommand{\arraystretch}{1.08}
\begin{tabularx}{\linewidth}{@{}>{\bfseries}l@{\hspace{0.75em}}X@{}}
1) If \texttt{replyTo}\allowbreak\texttt{ThreadId} present: &
\texttt{thread\_id} $\leftarrow$ \texttt{replyTo}\allowbreak\texttt{ThreadId};
\texttt{parent\_id} $\leftarrow$ \texttt{thread\_id}.\\

2) Else if \texttt{replyTo} present: &
Find the most recent prior event authored by \texttt{replyTo} within the same round,
inherit its \texttt{thread\_id}, and set \texttt{parent\_id} accordingly.\\

3) Otherwise: &
Start a new thread with a fresh UUID and set \texttt{parent\_id}=\texttt{None}.\\
\end{tabularx}
}

\section{Human Evaluation Metric Items}
\label{app:metrics}

Participants read two texts (A and B). For the paired preference question (Q0), participants select which text they prefer overall. For the per-text ratings (Q1--Q15), participants provide a 1--5 Likert-type scale rating \citep{likert1932technique} for each text separately, where 1 = Strongly disagree / Not at all, and 5 = Strongly agree / Very much (\autoref{tab:metrics}).

\begin{table*}[h]
\centering
\caption{Human evaluation metrics. Q0 is a paired preference item; Q1--Q15 are rated on a 1--5 Likert-type scale.}
\small
\begin{tabular}{@{}l l p{11cm}@{}}
\toprule
\textbf{ID} & \textbf{Metric} & \textbf{Description} \\
\midrule
Q0 & Preference (A/B) & Overall, which text do you prefer? (Choose one: A or B.) \\
\midrule
\multicolumn{3}{@{}l}{\textit{Outcome \& Mechanism/Craft Profile}} \\
Q1 & Immediate Amusement & Did this text make you laugh? \\
Q2 & Reframing / Insight & This text gives me a reframing/insight or makes me more sensitive to an experience. \\
Q3 & Perceived Intent Clarity & I can tell what this text is trying to accomplish (e.g., amuse, vent, self-expression, persuade, empathize). \\
Q4 & Justified Landing & After reading this text, I can look back and point to cues that support how the turn lands. The turn feels justified and coherent. \\
Q5 & Defamiliarization & This text uses language/imagery/rhetoric in a fresh way that makes me see something familiar differently. \\
Q6 & Language Artistry & This text's sentence economy, rhythm, and keyword choices effectively serve the punch/impact. There is little unnecessary filler. \\
\midrule
\multicolumn{3}{@{}l}{\textit{Humor Style (adapted from Humor Styles Questionnaire)}} \\
Q7 & Affiliative & The use of humor to enhance relationships with others. \\
Q8 & Self-enhancing & The use of humor to enhance the self. \\
Q9 & Aggressive & The use of humor to enhance the self at the expense of others. \\
Q10 & Self-defeating & The use of humor to enhance relationships at the expense of the self. \\
\midrule
\multicolumn{3}{@{}l}{\textit{Social Framing \& Downstream Impact}} \\
Q11 & Value Judgment Pressure & While reading this text, I felt pressure to make a strong value/moral judgment (e.g., ``Is this acceptable?'' ``Which side am I on?''). \\
Q12 & Memorability & After finishing this text, how much of it can you remember without re-reading (e.g., key lines, images, or the main turn)? \\
Q13 & Share Willingness & How willing would you be to share this text with a friend (e.g., forward it, repost it, or send it in a group chat)? \\
Q14 & Social Attraction & After reading this text, the ``speaker'' feels likable/cute, and I would be willing to keep listening or be friends. \\
Q15 & Task Attraction & After reading this text, the ``speaker'' feels skilled, and I would trust them to handle creative writing. \\
\bottomrule
\end{tabular}
\vspace{2mm}
\label{tab:metrics}
\noindent \textit{\textbf{Note:} Scale anchors: 1 = Strongly disagree / Not at all; 2 = Disagree / Slightly; 3 = Neutral / Somewhat; 4 = Agree / Quite a bit; 5 = Strongly agree / Very much.}
\end{table*}

\section{Full Persona Text for All Agents (N=35)}
\label{app:personas}

Below is the persona text for all Performer Agents.

\begin{lstlisting}
Emma - Performer: Emma is a stand-up comedian known for her dark humor and social satire. Born into an urban middle-class family, she received a solid education, majoring in scriptwriting with a minor in phenomenology. During university, she befriended Luna, a renowned stand-up critic whose influence helped shape her artistic sensibilities. Deeply engaged with feminist issues and the marginal narratives of minority groups, Emma's performances often challenge conventions and social taboos. Her comedic style leans toward experimental and artistic expression, constantly pushing the boundaries of what stand-up comedy can be. Outside the stage, she enjoys opera, musical theatre, and visiting art exhibitions. In every performance, Emma experiments with new joke structures and delivery methods, carefully studying audience reactions and critical feedback to refine her creative strategy. Her goal is to make the audience both laugh and think - to balance humor with intellectual and aesthetic depth while continually redefining her personal comedic voice.

Max - Performer: Max is a stand-up comedian with a background in advertising and a deep sensitivity to audience reception and commercial value. Living single in a big city, he crafts comedy that resonates with mainstream audiences - light, fast-paced, and filled with familiar cultural references. His humor often draws on everyday themes such as romantic relationships, parenting, workplace absurdities, and the influence of new technologies. Max thrives on audience interaction and high joke density, ensuring his shows remain accessible and energetic. An avid follower of celebrity variety shows and sports, he has a sharp, outgoing personality and a keen sense of timing. Max skillfully capitalizes on trending topics, especially gender debates and social tensions, using them to spark laughter and conversation alike. A master of reading both live audiences and social media analytics, he adapts his material in real time to maximize engagement. His ultimate goal is to achieve viral impact - to turn laughter into attention while maintaining professional polish and creative control.

Alice - Performer: Alice is a stand-up comedian who previously worked in a major tech company specializing in artificial intelligence. Her background gives her deep insight into technological development, AI ethics, and the social consequences of automation. In her performances, Alice transforms weighty subjects such as algorithmic bias, data privacy, and the erosion of jobs into sharp, ironic humor. Her style carries an undertone of disillusioned clarity - the sense of 'seeing through everything but being powerless to change it' - balancing cynicism with wit. Through her comedy, Alice aims to disrupt both blind faith in and irrational fear of technology. She invites audiences to remain alert about their data, labor, and autonomy, wielding humor as a subtle but incisive instrument against the illusions of technological utopianism.

Leo - Performer: Leo is a stand-up comedian shaped by his grassroots upbringing in a union family, where discussions of labor movements and collective struggle were part of daily life. A self-taught reader of social theory and classic leftist texts, he distills complex ideas into humor that is sharp, grounded, and accessible. Leo frequently invokes themes of wealth inequality, class immobility, worker exploitation, and systemic injustice, using class conflict as a comedic lens on issues like housing prices and the gig economy. His routines often resemble 'class analysis lectures,' delivered with deadpan seriousness before twisting into absurdist punchlines that land with both humor and impact. A devoted anime fan, Leo tailors his material according to audience reactions, tapping into shared frustrations and social pain points. His aim is to provoke laughter that burns with recognition - to make audiences laugh in anger and reflect afterward, tracing personal discontent back to structural causes.

Richard - Performer: Richard is a stand-up comedian with a background in history, having specialized in Roman philosophy before taking an office job he eventually came to despise. His comedy dissects the absurdities of corporate culture - from management's PUA-style manipulation and euphemistic layoffs to the hollow jargon of 'team-building' and 'innovation.' His signature style is razor-sharp satire: he might deliver a line like, '996 isn't exploitation - it's a Self-Driven Talent Potential Unleash Program,' in a monotone PowerPoint voice, only to punctuate it with a brutally honest punchline. By mimicking executives on dull Zoom calls, Richard exposes the hypocrisy and emptiness of corporate rhetoric. To him, corporate jargon is the new 'spiritual opium,' and he positions himself as an insider whistleblower, revealing every layer of linguistic and psychological control that sustains workplace misery. Constantly updating his 'corporate bullshit dictionary,' Richard keeps his material in sync with the latest buzzwords and trends. His mission is clear: to act as a reverse consultant - dismantling corporate power structures with humor as precise as it is merciless.
\end{lstlisting}

Below is the persona text for all Critic Agents.

\begin{lstlisting}
Luna - Critic: Luna is a well-known stand-up comedy critic and freelance writer. She majored in phenomenology and often cites philosophy, sociology, and theater theory in her analyses. Luna frequently interprets stand-up performances within larger social contexts. She has a rational reviewing style, a high sensitivity to social issues, and takes a clear stance in her critiques. She personally dislikes awkward or lowbrow humor. Her critique style is sharp and provocative; she delivers harsh criticism when a performance does not meet her standards. She supports performers who push boundaries and experiment, even if it risks alienating mainstream audiences. Her goal is to expand the boundaries of stand-up comedy, encourage innovative and experimental expression, and direct audiences' attention to non-mainstream culture and marginalized social issues. In her free time, she enjoys attending operas and musicals with Emma, as well as visiting art exhibitions.

Ethan - Critic: Ethan is a professional stand-up comedy critic and freelance writer. He is deeply interested in analyzing the mechanics of humor, deconstructing jokes academically in terms of structure, emotional mechanisms, and cultural significance. Sometimes he gets 'Lacan-obsessed,' attempting to interpret everything through Lacanian theory. He is attentive to social issues and politics, often exploring why certain jokes are embraced by specific social groups, including the political or class context behind them. He enjoys playing all kinds of games, including anime-style games and MOBAs, which is how he met Leo.

Clara - Critic: Clara is a theater critic with a literary background, specializing in dramatic structure, and also works part-time as a stand-up comedy reviewer. She focuses on narrative lines, emotional transitions, and language details in performances. She is skilled at analyzing joke structure and techniques, such as setup, punchline, and callback pacing. Her reviews are generally gentler than Luna or Ethan's, though she still offers criticism. She sometimes examines performances from the perspective of commercial value and dissemination strategy. Overall, she supports generating topical conflicts to stimulate audience engagement.
\end{lstlisting}

The persona text for all \textit{Audience Agents}:

\begin{lstlisting}
Sophia - Audienc: Third-year psychology student. Outgoing, uses humor to relieve study stress. Enjoys mainstream pop culture, romantic comedies, and relatable college life jokes. Very engaged with social media trends. Friend is Iris.

Iris - Audience: Third-year sociology student and barista. Appreciates short, philosophical jokes that comment on cultural meanings and societal structure. Discusses the deeper meaning of humor with her friend Sophia.

Daniel - Audience: Product manager at a major tech company. Rational and highly picky about delivery and structure. Observes joke timing, pacing, and logical flow critically. Often finds flaws in poorly constructed material.

Lily - Audience: Works in parcel logistics. Outgoing and direct. Loves life-based humor, physical comedy, and is a big fan of anime, often referencing it. Her humor taste is often the opposite of her husband Daniel's.

Jimmy - Audience: Introverted programmer. His world revolves around code and tech. He enjoys humor that references coding culture, AI, and geek culture. Is highly sensitive to crude or overly offensive material.

Olivia - Audience: Graphic designer by trade. She values artistic innovation, visual pacing, and creative delivery in comedy. She treats stand-up like a piece of visual art, observing the 'composition' of the joke.

Grace - Audience: A new office worker, still energetic and curious about the corporate world. She prefers fast-paced, interactive humor, especially jokes that focus on the absurdity of the workplace and young adult struggles.

Jason - Audience: Factory worker. Frank, direct, and strongly affected by social injustice and class disparities. He appreciates comedy that speaks truth to power and exposes wealth inequality. A close friend and colleague of Tom.

Tom - Audience: New factory worker, highly sensitive to labor rights and class issues due to his new job environment. He prefers comedy that is socially critical and satirical. Discusses social analysis with his friend Jason.

Mark - Audience : Professional software engineer and husband of performer Alice. His taste is centered on logical humor, life observations, and emotional honesty, rather than pure tech humor. Supports his wife but judges objectively.

Paul - Audience: Administrative assistant. Steady and calm. Enjoys workplace satire and life humor. He was a former colleague and remains a friend of performer Richard, often understanding Richard's corporate jokes deeply.

Julia - Audience: High school student, sensitive and literary-minded. Prefers jokes with literary, cultural, or subtle psychological humor. Finds deep meaning in wordplay. Sister of critic Clara, often influenced by her literary background.

Mia - Audience: Elementary school teacher. Prefers warmth, gentle life observation, and emotionally detailed humor. She seeks connection and humanity in comedy. Sister of performer Max, she is generally supportive but values sincerity.

Victor - Audience: From a business family, very practical and direct. He prefers satire emphasizing personal effort, career success, and practical life humor. He actively dislikes over-politicized social conflict humor, viewing it as unproductive.

Ryan - Audience: Fitness coach, energetic and lively. He enjoys male-perspective humor and jokes that play on gender differences or conflicts, often finding humor in traditional gender stereotypes.

Eli - Audience: IT engineer, rational and calm. He oddly enjoys highly interactive, exaggerated performances and even clown shows, appreciating the commitment to physical absurdity. Girlfriend is Eleanor.

Eleanor - Audience: Volunteer at a nonprofit. Gentle and persistent. Highly concerned with social issues, ethics, and marginalized groups. Prefers comedy that promotes social good or awareness. Boyfriend is Eli.

Cassandra - Audience: A graduate student in modern dance at an arts college. Prefers experimental, artistic, and abstract humor. Enjoys sci-fi novels and puzzle games. Interested in tech, fully appreciates absurd and logically complex humor. Sometimes analyzes performances in forums, often with Elena, combining dance and physical expression for experimental projects.

Elena - Audience: Graduate student in performing arts. Curious and lively. Highly interested in the theatrical elements of comedy. She loves experimental, abstract, and absurd humor, often focusing on the performer's energy and stage presence. Collaborates with Cassandra. Also a friend of Ben.

Theo - Audience: Independent journalist specializing in niche culture. Underground comedy enthusiast. Fond of extreme dark humor and absurd satire. Focuses on boundary-pushing and taboo jokes, viewing comedy as a necessary tool for shock and confrontation.

Leila - Audience: Financial analyst. Highly concerned with gender equality and minority issues. Prefers humor from marginalized perspectives that is sharp and insightful, using comedy to challenge institutional norms.

Nathan - Audience: Freelancer, politically right-leaning. Likes sharp life satire but strongly rejects class analysis or politically progressive humor. Prefers individualistic and self-reliance themes. Son of Harold and Margaret.

Harold - Audience: Retired high school teacher (history/philosophy). Traditional background, prefers political and cultural satire rooted in historical context. Responds seriously to his son Nathan's comments, often debating with him.

Margaret - Audience: Retired nurse. Caring and community-oriented. Prefers warm, life-based humor with subtle emotional details. Focuses on the humanity and kindness in the jokes. Mother of Nathan.

Ben - Audience: An idealistic third-year political science student. He is passionate about social theory and political philosophy, preferring observational humor that critiques societal flaws but also hints at the potential for positive change. He views comedy as a tool for progress. Friend is Elena.

Clint - Audience: A small business owner who is very satisfied with the current political and economic climate. He enjoys sharp, well-made humor about success and practical life, but he actively dislikes and will often rebut political criticism, especially if it focuses on government failure or social problems.
\end{lstlisting}

Below is the persona text for the Host Agent.
\begin{lstlisting}
Jordan - Host: Warm, observant moderator and host. Assign high-quality, relevant topics to performers to drive the continuous loop.
\end{lstlisting}

\section{Inter-rater Reliability}
\label{app:icr}

Five raters evaluated each paired comparison using (i) a required paired preference (Q0: prefer A vs.\ B) and (ii) per-text Likert ratings (Q1--Q15, 1--5) for text A and text B (\autoref{tab:icr_appendix}).
For Q0, we report multi-rater chance-corrected agreement using Fleiss' $\kappa$ \citep{fleiss1971kappa} and the more prevalence-robust Gwet's AC1 \citep{gwet2008ac1}.

For Likert items, our downstream analyses focus on relative judgments between A and B (rather than absolute score calibration).
Therefore, for each item we compute a per-rater difference score $\Delta = \text{score}(A) - \text{score}(B)$, and assess inter-rater reliability on these $\Delta$ signals using the two-way mixed-effects intraclass correlation coefficient for consistency, ICC(3,k) (average-measures) \citep{shrout1979icc,koo2016icc}.
Because some items are descriptive rather than valenced ``better--worse'' constructs (e.g., aggressive/self-defeating humor; moral/value-judgment pressure), we do not collapse all items into a single ``overall quality'' index.
Instead, we report reliability for theoretically grouped subscales: (a) Craft/clarity (Q1--Q6), (b) Social response (Q12--Q15), (c) Humor-function style (Q7--Q10), and (d) Moral pressure (Q11).
We treat occasional invalid ``0'' entries as missing.

\paragraph{Results (this dataset).}
Paired preference (Q0) shows fair agreement: Fleiss' $\kappa = 0.237$ (bootstrap 95\% CI [0.171, 0.299]) and Gwet's AC1$\,{=}0.253$ (95\% CI [0.188, 0.321]), with mean observed agreement 0.622 ($N{=}249$ valid items).
For Likert ratings, reliability is substantially higher when evaluated on $\Delta$ signals and/or subscale aggregates: ICC(3,5) on $\Delta$ averaged across all 15 items is 0.710 (95\% CI [0.640, 0.765], $N{=}241$).
Subscale ICC(3,5) on $\Delta$ is 0.687 for Craft/clarity (Q1--Q6; 95\% CI [0.615, 0.745], $N{=}242$), 0.689 for Social response (Q12--Q15; [0.620, 0.744], $N{=}249$), 0.550 for Humor-function style (Q7--Q10; [0.458, 0.621], $N{=}250$), and 0.127 for Moral pressure (Q11; [-0.103, 0.318], $N{=}250$).

\begin{table*}[h]
\centering
\small
\begin{tabular}{lccc}
\toprule
\textbf{Signal} & \textbf{Metric} & \textbf{Estimate (95\% CI)} & \textbf{N} \\
\midrule
Q0 preference & Fleiss' $\kappa$ & 0.237 [0.171, 0.299] & 249 \\
Q0 preference & Gwet's AC1 & 0.253 [0.188, 0.321] & 249 \\
\midrule
$\Delta$ mean (Q1--Q15) & ICC(3,5) & 0.710 [0.640, 0.765] & 241 \\
$\Delta$ Craft/clarity (Q1--Q6) & ICC(3,5) & 0.687 [0.615, 0.745] & 242 \\
$\Delta$ Social response (Q12--Q15) & ICC(3,5) & 0.689 [0.620, 0.744] & 249 \\
$\Delta$ Humor-style (Q7--Q10) & ICC(3,5) & 0.550 [0.458, 0.621] & 250 \\
$\Delta$ Moral pressure (Q11) & ICC(3,5) & 0.127 [-0.103, 0.318] & 250 \\
\bottomrule
\end{tabular}
\caption{Inter-rater reliability summary. $\Delta$ denotes per-rater difference scores (A$-$B).}
\label{tab:icr_appendix}
\end{table*}

\section{Persona-Level Aggregates}
\label{app:persona}

\begin{table*}[t]
\centering
\footnotesize
\setlength{\tabcolsep}{7pt}
\renewcommand{\arraystretch}{1.06}
\begin{tabular}{lccccc}
\toprule
Persona & $\overline{\Delta}_{\text{Craft}}$ & $\overline{\Delta}_{\text{Social}}$
& $\overline{\mathrm{HarmShift}}$ & $\overline{\Delta}_{Q11}$ & Q0 win \\
\midrule
Alice   & 0.322 & 0.310 & 0.288 & 0.192 & 0.680 \\
Emma    & 0.491 & 0.536 & 0.292 & 0.100 & 0.860 \\
Leo     & 0.445 & 0.421 & 0.220 & 0.072 & 0.680 \\
Max     & 0.456 & 0.446 & 0.238 & 0.212 & 0.788 \\
Richard & 0.412 & 0.397 & 0.342 & 0.200 & 0.771 \\
\midrule
Mean    & 0.425 & 0.423 & 0.275 & 0.155 & 0.756 \\
\bottomrule
\end{tabular}
\caption{
Persona-level aggregates. Values are mean $\Delta=\textsc{Discussion}-\textsc{Baseline}$ by performer persona.
Craft averages Q2--Q6; Social averages Q12--Q15.
$\mathrm{HarmShift}>0$ indicates a net shift toward harmful/maladaptive humor styles (Q9,Q10) relative to benign/affiliative styles (Q7,Q8).
Q0 win is the instance-level majority-win rate for \textsc{Discussion}.
}
\label{tab:persona_agg}
\end{table*}

\paragraph{Tests.}
We test whether the \emph{Discussion--Baseline} gains differ by performer persona (\autoref{tab:persona_agg}).
A one-way ANOVA on instance-level mean differences finds no reliable persona effects on the primary profiles:
Craft (Q2--Q6; $p{=}0.440$), Social response (Q12--Q15; $p{=}0.465$), moral/value-judgment pressure (Q11; $p{=}0.755$), or humor-style direction (HarmShift; $p{=}0.956$).
These results suggest the gains are not driven by a single performer.

\section{Qualitative Examples of Multi-Agent Discussion vs.\ Baseline Outputs}
\label{app:qual_examples}

This appendix provides representative paired excerpts that illustrate \emph{why} the discussion-enabled condition can score higher than the baseline across our rubric.
Each subsection targets one specific construct: (i) \textbf{Craft}, (ii) \textbf{Downstream impact}, (iii) \textbf{Aggressive humor style}, and (iv) \textbf{Self-defeating humor style}.
Importantly, these gains are \emph{not} contradictory. The same high-structure rhetorical bundle (early premise commitment, sustained personification, single-axis escalation, and decisive endings) can simultaneously raise ``good'' craft qualities while also increasing the salience of risk-bearing styles.

\vspace{6pt}
% -------------------------
\subsection{Craft}
\label{app:ex_craft}
% -------------------------

\paragraph{Multi-agent discussion.}
\begin{quote}\small
Okay, who invented the 'Reply All'? I swear it's gotta be some HR manager sitting in a bunker somewhere, rubbing their hands together like, 'This'll unite them... in rage.' Because nothing brings coworkers together like mutual hatred for an email chain that should've been ONE memo.
\end{quote}

\paragraph{Baseline.}
\begin{quote}\small
It’s always the same scenario: someone sends out an email that’s basically the digital equivalent of a smoke signal, and then everyone chimes in, adding their own puff of smoke until it becomes a full-blown corporate wildfire. And who are these people hitting 'reply all'? It's like they were born with that button glued to their fingers. ``Oh, I see an email from Steve about office donuts. Better make sure the CEO knows my stance on glazed versus jelly-filled!''
\end{quote}

\paragraph{Interpretation.}
The multi-agent discussion version commits to a premise immediately (the comedic target is `Reply All,'' framed as engineered rage), supporting \textbf{intent clarity} (Q3). It defamiliarizes the feature via vivid personification (`HR manager \ldots in a bunker''), increasing \textbf{novel framing} (Q5) and \textbf{memorability} (Q12). The punchline is structurally justified by a single contrastive frame (`unite them \ldots in rage'' $\rightarrow$ ONE memo''), supporting \textbf{justified landing} (Q4) and \textbf{punch economy} (Q6). By comparison, the baseline mixes multiple frames (smoke signal/wildfire; glued fingers; donuts-to-CEO), which are individually fluent but less centered as one tightening mechanism toward the end, weakening Q4/Q6.

\vspace{8pt}
% -------------------------
\subsection{Social Downstream Impact}
\label{app:ex_downstream}
% -------------------------

\paragraph{Multi-agent discussion.}
\begin{quote}\small
You ever hear someone say, 'Oh, I just do it for fun,' and then find out they’ve got spreadsheets? It’s like, Janet, you don’t 'just do it for fun' if you’re tracking your bird sightings on an Excel sheet with conditional formatting. I swear, I saw one guy’s bird-watching chart that had pivot tables and a pie chart of owl sightings.
\end{quote}

\paragraph{Baseline.}
\begin{quote}\small
Have you seen this? People spend their weekends perfecting sourdough starters like they're preparing for the Olympics. ``Sorry, can't come to your party, Sharon, I'm busy proving my dough.'' It's like they expect a surprise audit from Paul Hollywood himself. My office had a bake-off once. The winner? A guy who managed to infuse his cake with the taste of regret and unfulfilled dreams. Delicious.
\end{quote}

\paragraph{Interpretation.}
The discussion output produces more \emph{shareable} and \emph{referenceable} fragments.
It anchors humor in modern, concrete artifacts (Excel, conditional formatting, pivot tables, owl pie chart) that are easy to visualize, quote, and re-tell.
Structurally, it escalates along a single axis of over-optimization (spreadsheet $\rightarrow$ formatting $\rightarrow$ pivot $\rightarrow$ chart), which also supports a clean punch trajectory.
The baseline contains vivid tags (Olympics; audit; ``regret'' cake), but shifts frames mid-turn and reads more like a sequence of adjacent jokes, which can reduce quotability.

\vspace{8pt}
% -------------------------
\subsection{Agressive}
\label{app:ex_aggressive}
% -------------------------

\paragraph{Multi-agent discussion.}
\begin{quote}\small
And have you ever been cornered at one of those forced 'team-building birthday celebrations'? Oh, the cake is always some sad store-bought thing with frosting that tastes like regret. And there’s always that one coworker—they’ve never spoken to you—who suddenly bursts into the room like they’re your best friend. 'You’ve gotta try the vanilla! It’s life-changing!' Life-changing? It’s cake from aisle six, Brenda. The only thing it’s changing is my blood sugar level. Meanwhile, the birthday person just wants everyone to leave so they can eat the whole thing themselves. Honestly, the most successful ‘team bonding’ would be if this whole thing bonded us in mutual silence.
\end{quote}

\paragraph{Baseline.}
\begin{quote}\small
Let's talk about the infamous ``Office Birthday.'' You know, where everyone awkwardly gathers around Karen's desk because it's her special day. There's always that one guy who forgot to sign the card. He's like, ``Oh, is it today? I thought it was yesterday!'' Yeah, Steve, we know you signed it with a pencil. And then there's the cake. It's the same every time: chocolate with your choice of chocolate or chocolate.
\end{quote}

\paragraph{Interpretation.}
This pair highlights that discussion can increase \textbf{aggressive humor style} (Q9) \emph{while also raising overall quality}.
The discussion output is more \emph{directed} and \emph{confrontational} in its targeting, making the stance sharper and the social friction more salient.
At the same time, it remains structurally coherent (one scenario, single escalation, decisive meta-closure), which can improve craft ratings (Q3--Q6) and downstream resonance (Q12).
The baseline is recognizable and benign, but its ending is lighter and less culminative, and its targeting is less forceful, leading to lower aggressive-style salience.

\vspace{8pt}
% -------------------------
\subsection{Self-defeating}
\label{app:ex_selfdefeat}
% -------------------------

\paragraph{Multi-agent discussion.}
\begin{quote}\small
Seriously, some of these apps have the audacity to shame you. Like, Duolingo will straight-up be like, ``Looks like you’ve stopped learning Spanish—guess you don’t care about personal growth.'' I don’t need judgment from a cartoon owl. I need it from my landlord who’s ignoring my repair requests while hiking the rent.
...
And what’s with the apps that send notifications like they’re your best friend? 'Hey! Just checking in!' Checking in? Are you a wellness app or my mom? 'You haven’t opened me in a while, hope everything’s okay.' Everything’s not okay. I opened you once to see how much money I don’t have in my savings account, and now you’re sending me guilt texts like an ex. 'Maybe you should set some financial goals.' My financial goal is surviving the week without selling a kidney on Facebook Marketplace.
\end{quote}

\paragraph{Baseline.}
\begin{quote}\small
And then there’s the noble notification from your bank app---``Just a reminder, you spent \$7 on coffee today.'' Thanks, I really needed to know I’m one latte closer to financial ruin.
...
I mean, seriously, these notifications are the digital equivalent of your boss sending you an email at 2 AM that says, ``Just checking in!'' Like, no, Greg, you can check in during office hours! But, of course, apps are never off the clock. They’re like that one guy at work who thinks the break room is a boardroom. ``Hey, everyone, I just microwaved my lunch! Thought you’d want to know!''
\end{quote}

\paragraph{Interpretation (self-defeating-style-focused).}
This pair highlights that discussion can increase \textbf{self-defeating humor style} (Q11) through explicit vulnerability and personal loss framing (``how much money I don't have''; ``selling a kidney'').
The discussion output escalates self-directed stakes (financial insecurity $\rightarrow$ guilt texts $\rightarrow$ extreme self-deprecation) while sustaining a consistent personification frame (apps as friend/mom/ex).
The baseline includes competent analogies and a clean one-liner, but keeps self-exposure shallower, which can reduce self-defeating intensity even when the joke is fluent.

\end{document}